\documentclass{elsart}
\usepackage[english]{babel}
\usepackage{amssymb}
\usepackage{amsfonts}
\usepackage{amsmath}
\usepackage{natbib}
\usepackage{url}
\usepackage{fancyhdr}
\usepackage{graphicx}
\usepackage{calc}

\newcommand{\R}[1]{\ensuremath{\Rset^{#1}}}

\newcommand{\splineorder}{\nu}
\newcommand{\LeTitre}{Representation of Functional Data in Neural Networks}

\usepackage[bookmarks=true,ps2pdf=true,hyperindex=true,colorlinks=false,
pdftitle={Representation of Functional Data in Neural Networks},
pdfkeywords={Functional data analysis, Multi-Layer Perceptron, Radial Basis Function Network, Supervised learning, Curves discrimination, Nonlinear functional model, Spectrometric data},
pdfsubject={Functional Data Analysis with Neural Networks},
pdfauthor={Fabrice Rossi, Nicolas Delannay, Brieuc Conan-Guez, and Michel Verleysen}]{hyperref} 
\begin{document}
\date{26 October 2004}
\begin{frontmatter}
\title{\LeTitre\protect\footnote{Published in Neurocomputing (Volume 64,
  pages 183--210). DOI: \url{http://dx.doi.org/10.1016/j.neucom.2004.11.012}}}
\author[INRIA,CEREMADE]{Fabrice Rossi\corauthref{rossi}},
\ead{Fabrice.Rossi@inria.fr}
\author[DICE]{Nicolas Delannay\thanksref{FNRSRF}},
\ead{delannay@dice.ucl.ac.be}
\author[INRIA]{Brieuc Conan-Guez}
\ead{Brieuc.Conan-Guez@inria.fr}
and
\author[DICE]{Michel Verleysen\thanksref{FNRSSR}}
\ead{verleysen@dice.ucl.ac.be}
\address[INRIA]{Projet AxIS, INRIA-Rocquencourt, Domaine de Voluceau,
  Rocquencourt, B.P.~105,
78153 Le Chesnay Cedex, France}
\address[CEREMADE]{CEREMADE, UMR CNRS 7534, Universit\'e Paris-IX Dauphine,
Place~du~Mar\'echal de Lattre de Tassigny, 75016 Paris, France}
\address[DICE]{Universit\'e catholique de Louvain (UCL)
        DICE - Machine Learning Group
        Place~du levant 3, 1348 Louvain-la-Neuve, Belgium }
\corauth[rossi]{Corresponding author:\\
Fabrice Rossi\\
Projet AxIS\\
INRIA Rocquencourt\\
Domaine de Voluceau, Rocquencourt, B.P. 105\\
78153 LE CHESNAY CEDEX -- FRANCE\\
Tel: (33) 1 39 63 54 45\\
Fax: (33) 1 39 63 58 92\\
}
\thanks[FNRSRF]{Nicolas Delannay is Research Fellow of the Belgian National
  Fund for Scientific Research.} 
\thanks[FNRSSR]{Michel Verleysen is Senior Research Associate of the Belgian
  National Fund for Scientific Research.}
\end{frontmatter}

\newpage

\begin{frontmatter}
\title{\LeTitre}
\begin{abstract}
  Functional Data Analysis (FDA) is an extension of traditional data analysis
  to functional data, for example spectra, temporal series, spatio-temporal
  images, gesture recognition data, etc.  Functional data are rarely known in
  practice; usually a regular or irregular sampling is known.  For this
  reason, some processing is needed in order to benefit from the smooth
  character of functional data in the analysis methods.  This paper shows how
  to extend the Radial-Basis Function Networks (RBFN) and Multi-Layer
  Perceptron (MLP) models to functional data inputs, in particular when the
  latter are known through lists of input-output pairs.  Various possibilities
  for functional processing are discussed, including the projection on smooth
  bases, Functional Principal Component Analysis, functional centering and
  reduction, and the use of differential operators.  It is shown how to
  incorporate these functional processing into the RBFN and MLP models. The
  functional approach is illustrated on a benchmark of spectrometric data
  analysis. 
\end{abstract}
\end{frontmatter}

\newpage

\section{Introduction}
Many modern measurement devices are able to produce high-resolution data
resulting in high-dimensional input vectors. A promising way to handle this
type of data is to make explicit use of their internal structure. Indeed,
high-resolution data can frequently be identified as discretized functions:
this is the case for time series (in the time domain as well as in the
frequency domain), spectrometric data, weather data (in which we can have both
time and location dependencies), etc. Functional Data Analysis (FDA, see
\cite{RamseySilverman97}) is an extension of traditional data analysis methods
to this kind of data. In FDA, each individual is characterized by one or more
real-valued functions, rather than by a vector of \R{p}. Function estimates
are constructed from high-dimensional observation vectors and data analysis
(in a broad sense) is carried out on those estimates.

As it is not possible to directly manipulate arbitrary functions, a
computer-friendly representation of functional data must be used: this is
obtained through a basis expansion in the functional space, for instance with
a B-spline approximation. This way of proceeding has numerous advantages over
a basic multivariate analysis of high-dimensional data. Indeed the choice of a
fixed basis allows to introduce prior knowledge: for instance a Fourier basis
can be used to model periodic functions such as daily temperature observations
in a fixed location.  A fixed basis also allows to deal with irregularly
sampled functions and with missing data. A side effect of the representation
is that it can be used to smooth the data either individually or globally (see
\cite{BesseCardotFerraty1997}). Another interesting point is that most FDA
methods can work directly on the numerical coefficients of the basis
expansion, leading to far less computational burden.  An additional advantage
of dealing with functions is the possibility of using functional preprocessing
such as derivation, integration, etc.

FDA is based on the fact that the first step performed by many data analysis
methods consists in simple operations on the data: distance, scalar
product and linear combination calculations. Those operations can be defined
in a satisfactory way in arbitrary Hilbert spaces that include functional
spaces (such as $L^2$). This means that many data analysis methods can be
extended to work directly with functional inputs. There are of course some
theoretical difficulties induced by the infinite dimension of the considered
spaces.  Nevertheless, traditional data analysis methods have been
successfully adapted to functional data, both on theoretical and practical
point of views. We refer to \cite{RamseySilverman97} for a comprehensive
introduction to those methods, especially to functional principal component
analysis and functional linear models. For regression and discrimination
problems, recent developments of FDA include nonlinear models such as
multilayer perceptrons
\cite{RossiConanFleuretFMLPESSAN2002,RossiConanFleuretFMLPWCCI2002},
semi-parametric \cite{FerreYao2003SIR} and non-parametric models
\cite{FerratyVieu2002CS,FerratyVieu2002CSDA}.

In this paper, we extend the results from
\cite{DelannayEtAlESANN2004RbfFunc,RossiConanESANN2004FMLPMissing}, and
introduce two nonlinear neural models for functional data. They are adaptation
of classical neural models, the Radial-Basis Function Network (RBFN) and the
Multi-Layer Perceptron (MLP), to functional inputs.  In particular, we show how
to implement these functional models in practical situations, when the
functional data are known through a list of numerical samples (regularly or
irregularly sampled input-output pairs).  In section
\ref{sectionFunctionalSpaces} we first show that many data analysis algorithms
can be defined on arbitrary Hilbert spaces, that include the $L^2$ functional
space with its elementary operations. We illustrate this general construction
with the proposed models, using distances in $L^2$ for the RBFN and inner
products for the MLP. Section \ref{sectionFunctionRepresentation} introduces
the general FDA method that uses function representation to actually implement
theoretical models defined in $L^2$. We show that the MLP and RBFN models can
be implemented on preprocessed coordinates of the functional data,
providing a way to easily introduce functional data processing in classical
neural software. We also introduce in this section natural preprocessings that
are available for functional data, such as functional principal component
analysis, derivation, etc. In section \ref{sectionSimulations} we report
simulation results on a real-world benchmark, a spectrometric problem in which
the percentage of fat contained in a meat sample must be predicted based on
its near-infrared spectrum. We show that functional preprocessing greatly
improves the performances of the RBFN and gives very good performances with a
simple MLP. In section \ref{sectionMissingData}, we build a semi-artificial
dataset, introducing randomly placed holes in the spectrum data. This
simulates irregular sampling in its simplest form (missing data). We show that
the functional reconstruction allows to maintain excellent predictions whereas
classical data imputation techniques are not able to reconstruct the missing
information.

\section{Working directly in functional spaces}\label{sectionFunctionalSpaces}
\subsection{Introduction}
In this paper, we focus on regular functions, i.e. on square integrable
functions from $V$, a compact subset of \R{p}, to \R{}. We denote $L^2(V)$ the
vector space of those functions. A $L^2(V)$ space equipped with its natural
inner product $\langle f,g\rangle=\int_V f(x)g(x)\d x$, is a Hilbert vector
space. In the present section, we will avoid using specific aspects of
$L^2(V)$. We will rather illustrate how elementary operations available in a
Hilbert space as linear combinations, inner product, norm and distance
calculations are sufficient to implement many data analysis algorithms, at
least on a theoretical point of view.

In this section, $H$ denotes an arbitrary Hilbert space. When $u$ and $v$ are
arbitrary elements of $H$, $\langle u,v\rangle$ denotes their inner product,
and $\|u\|=\sqrt{\langle u,u\rangle}$ the norm of $u$.

\subsection{Data analysis in a Hilbert space}
Even if data analysis algorithms have been defined for traditional
multivariate observations, they seldom use explicitely the finite dimensional
character of the input spaces. The most obvious cases are distance-based
algorithms such as the $k$-means method.

Indeed, the $k$-means algorithm clusters input data by alternating between two
phases: an affectation phase and a representation phase. The goal is to obtain
representative clusters; each of them is defined by a prototype that
belongs to the input space. Given the prototypes, the
\emph{affectation} phase puts input vector $x$ in the cluster defined by the
prototype closest to $x$: obviously we only need to calculate distances
between points in the input space to perform this affectation operation. The
\emph{representation} phase consists in updating the prototypes given by the
results of the affectation phase. For a given cluster, the new prototype is
defined as the center of gravity of the input vectors associated to the
considered cluster. The new prototype is therefore a linear combination of
input vectors. Hence $k$-means can be defined for any normed vectorial
input space (an inner product is not even needed); this obviously includes
functional spaces. The $k$-means method has been adapted to $L^2$
spaces in \cite{Abraham2000} in which the consistency of the algorithm is
proved (see also \cite{JamesSugar2002} for a EM-like version of a functional
clustering algorithm).

More sophisticated clustering methods such as the Self-Organizing Map (SOM,
\cite{KohonenSOM1997}) are also based on elementary operations (distance and
linear combination calculations). They can therefore be applied to functional
input spaces (see \cite{RossiConanGuezElGolliESANN2004SOMFunc} for the SOM
applied to functional data).

Regression models can also be constructed when the explanatory variable
belongs to arbitrary normed vector spaces. Let us consider for instance the
linear regression: the goal is to model a random variable $Y$ (the target
variable with values in \R{}) as a linear function of a random vector $X$ (the
input variable), i.e. $E(Y|X)=l(X)$. If $X$ has values in \R{p}, an explicit
numerical representation of the linear function can be written such as
$l(X)=\sum_{i=1}^p\alpha_iX_i$, where $X_i$ is the $i$-th coordinate of $X$.
More generally if $X$ has values in an arbitrary normed vector space $M$, it
is still possible to model $Y$ by $E(Y|X)=l(X)$ by requesting $l$ to belong to
$M^*$, the topological dual of $M$, i.e. the set of continuous linear
functions from $M$ to \R{}. In the particular case of a Hilbert space $H$, the
identification of $H$ with its dual $H^*$ is used to obtain a simpler
formulation. More precisely, any continuous linear form $l$ on $H$ can be
represented through an element $v\in H$ such that $l(u)=\langle u,v\rangle$.
We have therefore $E(Y|X)=\langle X,v\rangle$ for a well chosen $v\in H$.

Of course, the linear model is a very limited regression model. Non-parametric
models are a possible solution to overcome those limitations: they have been
extended to functional data in \cite{FerratyVieu2002CS,FerratyVieu2002CSDA}.
Semi-parametric models have also been adapted to functional inputs in
\cite{FerreYao2003SIR}. We propose in this paper to build neural network based
nonlinear regression models. In the following sections, we show how to define
Radial-Basis Function Networks (RBFNs) and Multi-Layer Perceptrons (MLPs) with
functional inputs.

\subsection{Radial-Basis Function Networks}\label{sectionRBFN}

Radial-Basis Function Networks (RBFN) are popular nonlinear models that have
several advantages over other nonlinear regression paradigms. Besides their
simplicity, their intuitive formulation and their local approximation
abilities, their most important advantage is probably the ability of various
learning procedures to avoid the local minima issue, for example when the
parameters of the model are the solution of a linear problem.

The first operation performed by RBFN models on the input data is based on the
notion of distances. The following of this section shows how this notion can
be inserted in a functional data context and more generally in any metric
space such as a Hilbert space. 

The output of a RBF network is expressed by
\begin{equation}
y=\sum^{p}_{i=1}\alpha_{i}\varphi_i\left(d_i(x,c_i)\right),
\label{ModelRBFN}
\end{equation}
where $x$ is the input of the network, $y$ its scalar output,
$\varphi_i(\cdot)$ are radial-basis functions from $\R{}$ to $\R{}$, $c_i$ are
centers chosen in the input data space of $x$, $d_i(\cdot,\cdot)$ are
associated distances and $\alpha_{i}$ are weighting coefficients.  We see that
the predicted output is expressed as a weighted sum of basis functions with
radial shape (each basis function has a radial symmetry around a center). This
property is very general; any distance measure between the input $x$ and the
centers $c_i$ could be used. Most frequently the RBF are Gaussian
$\varphi_i(r)= \exp(-r^2)$.

Equation \ref{ModelRBFN} easily generalizes to any metric space by replacing
all distances $d_i$ by the distance used to define the space. In the
particular case of a functional space, distance $d_i(x,c_i)$ between vectors
is simply replaced by a distance $d_i(g(\cdot),c_i(\cdot))$ between the
functional input $g(\cdot)$ and the functional centers $c_i(\cdot)$.

Some RBF (e.g. Gaussian functions) can define the positive definite kernel of
a Reproducing Kernel Hilbert Space (RKHS) $k(x,x') = \varphi(d(x,x'))$.
Regularization networks \cite{cit_Girosi_RegTheo} and Support Vector Machines
(SVM) result from a learning theory within this RKHS context
\cite{cit_SLTbook,cit_Evgeniou_RNSVM}.  Therefore the following discussion
straightforwardly applies to these models too.

RBF networks have the universal approximation
property~\cite{cit_Park_UnivApprox}. In practice, when a finite number of
observed data is available, the way the distance is defined plays a crucial
role on the generalization performances of the network. Dealing with
functional data, the kind of distance to be used must be specified. For
example, considering functions from $L^2(V)$, the Euclidean distance
in this space could be chosen:
\begin{equation}
d_i(g(\cdot),c_i(\cdot))={\left(\int_{V}(g(x)-c_i(x))^{2}dx\right)}^{1/2}.
\label{FuEuclDist}
\end{equation}
While this choice might reveal adequate in some situations, the Euclidean
distance is in fact quite restrictive.  In the vectorial case, one could use
a weighted distance (for example the Mahalanobis one) instead of the Euclidean distance; generalizing to
functional spaces, a weighted version of the Euclidean distance could be used
to characterize the measure of locality around each center:
\begin{equation}
d_i(g(\cdot),c_i(\cdot))={\left(\int_V\int_V(g(x')-c_i(x'))w_i(x',x)(g(x)-c_i(x))dxdx'\right)}^{1/2},
\label{FuWeightEuclDist}
\end{equation}
where $w_i(x',x)$ is a positive definite bivariate function over $V \times V$.
While this last definition is very general, there is unfortunately no simple
way to choose the weighting function $w_i(x',x)$. It seems reasonable to look
for a weighting function that shows approximately the same complexity as the
input data (the complexity of a function being often measured through its
second derivative). Using a smooth weighting function leads to the so-called
functional regularization. In other words, working with functional data
necessitates a functional regularization of the parameters defining the
distance measure.

Another possibility resulting from the use of functional data is that
differential operators can be applied.  This could reveal interesting for
example when the shape of the functional inputs is known to be more important
than their absolute levels (or means); see section \ref{sectionRBFResults} for
an application example. In the framework of RBF networks, differential
operators can also be included in the distance function, which becomes a
semi-metric:
\begin{equation}
d_i(g(\cdot),c_i(\cdot))={\left(\int_{V}\left(Dg(x)-c_i(x)\right)^{2}dx\right)}^{1/2},
\label{FuDist2}
\end{equation}
where $\mathit{D}(\cdot)$ is a differential operator (for example the first or
second derivative, as used in sections \ref{sectionSimulations} and
\ref{sectionMissingData}). 

\subsection{Multi-layer Perceptrons}
A multilayer perceptron (MLP) consists in neurons that perform
very simple calculations. Given an input $x\in\R{p}$, the output of a
neuron is 
\begin{equation}
T(\beta_0+\sum_{i=1}^p\beta_ix_i), 
\end{equation}
where $x_i$ is the $i$-th coordinate of $x$, $T$ is a nonlinear activation
function from \R{} to \R{}, and $\beta_0,\ldots,\beta_p$ are numerical
parameters (the weights of the neuron).

As for the linear model considered previously, this calculation can be
generalized to any normed vector space $M$ (see
\cite{Sandberg1996,SandbergXu1996,Stinchcombe99}). If $l$ is a linear form in
$M^*$, it can be used to define a neuron with an input in $M$ and whose output
is given by $T(\beta_0+l(x))$ for $x\in M$. The linear form replaces parameters
$\beta_1,\ldots,\beta_p$. Obviously, the case $M=\R{p}$ corresponds
exactly to the traditional numerical neuron.

In a Hilbert space $H$, linear forms are represented by inner products and
define a generalized neuron with an input in $H$: given an input vector $u$,
the neuron output is $T(\beta_0+\langle u,v\rangle)$. The ``connection
weights'' of the neuron are the numerical value $\beta_0$ and the vector
$v\in H$. In the particular case of $H=L^2(V)$, given an input function $g$,
the neuron output is $T(\beta_0+\int_V g(x)w(x)\d x)$. The neuron is called a
functional neuron and $w$ is its \emph{weight function}.

As the output of a generalized neuron is a numerical value, we need such
neurons only in the first layer of the MLP. Indeed, the second layer uses only
outputs from the first layer which are real numbers and therefore consists in
numerical neurons.

We have presented in \cite{RossiConanFleuretFMLPESSAN2002} and
\cite{RossiConanGuez04NeuralNetworks} some theoretical properties of
MLPs constructed by combining a layer of generalized neurons with inputs in
$L^2(V)$ and at least one layer of numerical neurons. We use specific
properties of $L^2(V)$ that allow to restrict the set of ``connection
weights'': rather than working with arbitrary weight functions in $L^2(V)$, we
use weight functions that can be exactly calculated by a traditional MLP or by
any other sufficiently powerful function approximation method. An important
result is that MLPs with functional inputs are universal approximators as long
as they use sufficiently regular activation functions, exactly as numerical
MLPs: given a continuous function $G$ from $K$ a compact subset of $L^2(V)$ to
\R{} and $\epsilon>0$ an arbitrary positive real number, there is an
one-hidden layer perceptron that calculates a function $H$ such that
$|G(g)-H(g)|<\epsilon$ for all $g\in K$. This MLP uses functional neurons in
its first (hidden) layer and one numerical neuron (with the identity
activation function) in its output layer. We have also shown in
\cite{ConanRossiICANN2002} that even if the MLP is implemented through a
function representation (as it will be described in the following section),
the universal approximation property is still valid.

\section{Function representation}\label{sectionFunctionRepresentation}
\subsection{Functional data in practice}
The previous section shows that it is possible to define many data analysis
algorithms for arbitrary Hilbert spaces. However, the proposed solutions are
purely theoretical; it is in general impossible to manipulate arbitrary
functions from $L^2(V)$ on a computer. Moreover, functional data coming from
sensors, measurements or collected in other ways do not consist in
mathematical functions. On the contrary, as stated in the introduction,
observations are discretized functions: each of them is a list of input/output
pairs. These lists may include missing data or more generally show irregular
sampling: the sets of inputs for each observation do not necessarily coincide.

More precisely, let us assume that we observe $n$ functions such that function
$i$ is given by the $(x^i_j,y^i_j)_{1\leq j\leq m^i}$ list of $m^i$ pairs,
with $x^i_j\in V$ and $y^i_j\in\R{}$. FDA main assumption is that there is a
regular function $g^i$ (in $L^2(V)$) such that
$y^i_j=g^i(x^i_j)+\epsilon^i_j$, where $\epsilon^i_j$ is an observation noise.
In this model, both the number of observations $m^i$ and the $(x^i_j)_{1\leq
  j\leq m^i}$ can depend on $i$.

The $g^i$ functions are not known. This prohibits the straightforward
application of the models developed in the previous section. Even with known
functions, calculating elementary operations, such as integrals, is difficult.
Nevertheless the rationale of FDA is to implement theoretical models on
those functions. A possible solution, quite common in FDA methods, is to
construct an approximation of the $(g^i)_{1\leq i\leq n}$ and then to work on
these approximations. One way to build them is to project
the original $g^i$ on a known subspace.

\subsection{Representation on a subspace}\label{subsectionSubSpace}
FDA introduces some specific needs that have to be taken into account to
choose a representation subspace. A first need is that the representation must be computed for every input list; this computation should therefore be as fast as possible.  A
second constraint is that the operations performed on the reconstructed
functions must approximate as exactly as possible the corresponding operations
on the original $g^i$ functions.

\subsubsection{An approximate projection}
A simple and efficient solution is provided by a projection approach which
makes use of the $L^2(V)$ Hilbert structure. A set of $q$ linearly independent
functions from $L^2(V)$, $(\phi_k)_{1\leq k\leq q}$ is chosen. Rather than
working on $L^2(V)$, we restrict ourselves to
$\mathcal{A}=span(\phi_1,\ldots,\phi_q)$ and use $(\phi_k)_{1\leq k\leq q}$ as
a basis for this subspace.  Each function $u\in \mathcal{A}$ is represented by
its $\alpha_k(u)$ coordinates, such that
$u=\sum_{k=1}^q\alpha_k(u)\phi_k$.

Given a list $(x^i_j,y^i_j)_{1\leq j\leq m^i}$, the underlying function $g^i$
is then approximated by a function $\tilde{g}^i$ in $\mathcal{A}$. The best
approximation in the functional sense would be to choose $\tilde{g}^i$ as the
orthogonal projection of $g^i$ on $\mathcal{A}$. Obviously, such projection
cannot be calculated exactly as we do not know $g^i$.  Therefore,
$\tilde{g}^i$ is defined by its numerical coefficients
$(\alpha_k(\tilde{g}^i))_{1\leq k\leq q}$ chosen to minimize:
\begin{equation}\label{eqDistortionBasis}
\sum_{j=1}^{m^i}\left(y^i_j-\sum_{k=1}^q\alpha_k(\tilde{g}^i)\phi_k(x^i_j)\right)^2.
\end{equation}
This minimization is a standard quadratic optimization problem that
can be conducted very efficiently with cost at most $O(m^iq^2)$ (see
\cite{RamseySilverman97} chapter 3 for instance). Moreover, some specific
functional bases such as B-splines lead to even faster algorithms with cost
$O(m^iq)$ in some situations.

In the next subsection, it is shown that this representation
approach also allows to transform the functional operations on
$\tilde{g}^i$ into calculations on the
$\alpha_k(\tilde{g}^i)$ coordinates.

\subsubsection{Working with the coefficients}\label{sectionCoeff}
As $\mathcal{A}$ is a finite-dimensional space, it is possible to
work with the coordinates $\alpha_k(\tilde{g}^i)$ instead of working
directly on the $\tilde{g}^i$ functions.  Nevertheless, it is shown
in the following that additional precautions must be taken if the
basis functions $(\phi_k)_{1\leq k\leq q}$ are not orthonormal.

Once each functional input data is transformed into a vector in
$\R{q}$ that corresponds to its coordinates in $\mathcal{A}$,
traditional data analysis algorithms can be used directly on those
vectors.  However, while this simple approach can give good results
in some situations, it introduces an unwanted distortion in the
input function representation.

The case of linear operations does not introduce any problem. Indeed a linear
combination of functions may be expressed as a linear combination of their
coordinate vectors: if $u$ and $v$ are functions in $\mathcal{A}$ and
$\lambda$ and $\mu$ are real numbers, then $\alpha_k(\lambda u+\mu
v)=\lambda\alpha_k(u)+\mu\alpha_k(v)$ for all $k$.

On the contrary, inner products, and therefore distances, are a source of
problems. Indeed the inner product between two functions in $\mathcal{A}$ can
also be defined in terms of their coordinate vectors. The inner product
between $u$ and $v$ in $\mathcal{A}$ is given by:
\[
\langle u,v\rangle = \sum_{k=1}^q\sum_{l=1}^q\alpha_k(u)\alpha_l(v)
\langle \phi_k,\phi_l\rangle
\]
If we denote $\alpha(u)=(\alpha_1(u),\ldots,\alpha_q(u))^T$, where
${^T}$ is the transposition operator, the inner product becomes:
\[
\langle u,v\rangle = \alpha(u)^T\Phi\alpha(v),
\]
where $\Phi$ is the matrix defined by $\Phi_{kl}=\langle
\phi_k,\phi_l\rangle$ and independent from $u$ and $v$.

This last formula shows that a distortion corresponding to $\Phi$ results from
the transition between the inner product $\langle u,v\rangle$ in $\mathcal{A}$
and the canonical inner product in $\R{q}$. The norm of the difference between
functions of course shows the same behavior: in general,
$\|\alpha(u)-\alpha(v)\|^2$ is different from $\|u-v\|^2$, as the latter is
given by $(\alpha(u)-\alpha(v))^T\Phi(\alpha(u)-\alpha(v))$, while the former
is simply $(\alpha(u)-\alpha(v))^T(\alpha(u)-\alpha(v))$. If the set of
functions $(\phi_k)_{1\leq k\leq q}$ is not orthonormal, $\Phi$ is different
from the identity matrix and the inner product that should be used in $\R{q}$
is not the canonical one. Unfortunately, some very useful sets of functions,
such the B-splines (see section \ref{sectionBasis}), are not orthonormal.

A simple solution to this problem is to use the Choleski
decomposition of $\Phi$, i.e. a square matrix $U$ such that
$\Phi=U^TU$.  The coordinate vectors can then be scaled by matrix
$U$ to give $\beta(u)=U\alpha(u)$. Obviously, we have:
\[
\beta(\lambda u+\mu v)=\lambda\beta(u)+\mu\beta(v)
\]
and
\[
\beta(u)^T\beta(v)=\sum_{k=1}^q\beta_k(u)\beta_k(v)=\langle
u,v\rangle.
\]
This means that performing elementary operations in $\R{q}$
(with its canonical inner product) on the coordinates $\beta(u)$ is
exactly equivalent to performing the same operations in the inner
product space $\mathcal{A}$.

Working with the coordinate vectors $\beta(\tilde{g}^i)$ is thus strictly
equivalent to working directly on the $\tilde{g}^i$ functions, and equivalent
to working with $g^i$ under the approximation resulting from equation
\ref{eqDistortionBasis}. A nice consequence of this property is that
functional models can be implemented as a preprocessing phase before any
classical data analysis software: the preprocessing consists in choosing the
projection space (see the following section) and in calculating the coordinate
vectors $\beta(\tilde{g}^i)$. Then, those vectors can be submitted to a RBFN
or a MLP exactly as classical multivariate data. Optionally, some additional
functional preprocessing can be implemented before the final transformation
(see section \ref{sectionTransformation}). In general, the calculation of
$\Phi$ does not introduce any additional problems as the basis functions are
under the practitioner control. With an orthonormal basis such as Fourier
series, $\Phi$ is the identity matrix. For other bases, quadrature methods or
Monte-Carlo methods can be used to calculate an arbitrarily accurate
approximation of $\Phi$.

\subsection{Choosing the projection space}
As the underlying functions are reconstructed according to their approximation
in the projection space, the choice of this space has an important impact on
the data analysis. The projection space must for instance provide a good
approximation of arbitrary functions in $L^2(V)$ as there is no \emph{a
  priori} reason to restrict the functions to a subspace of $L^2(V)$.

\subsubsection{Basis}\label{sectionBasis}
Good candidate bases are provided by Hilbert bases of $L^2(V)$. Given such a
basis $(\phi_k)_{1\leq k}$, truncation allows to define finite-dimensional
subspaces: $\mathcal{A}_q=span(\phi_1,\ldots,\phi_q)$. An interesting
theoretical property of Hilbert bases is that functions from $L^2(V)$ can be
approximated more and more accurately by increasing the number $q$ of basis
functions. In practice, when the Hilbert basis is fixed, a leave-one-out
technique allows to choose $q$ directly from the data (see section
\ref{sectionCrossValidation}). An example of Hilbert basis is given by the
Fourier series for $L^2([a,b])$, where $[a,b]$ is an interval in \R{}.

Another interesting solution is provided by B-splines. Let us assume that
$V=[a,b]$ and let $\pi=(t_0,t_1,\ldots,t_{l+1})$ be a sequence such that
$t_0=a$, $t_{l+1}=b$ and $t_k<t_{k+1}$ for all $k$. With $\splineorder$ a
positive integer, we denote $\mathcal{S}^{\splineorder}_{\pi}$ the subset of
$L^2([a,b])$ defined as follows: a function $f\in L^2([a,b])$ belongs to
$\mathcal{S}^{\splineorder}_{\pi}$ if $f$ is $C^{\splineorder-2}$ on $[a,b]$
and if $f$ is a polynomial of degree $\splineorder-1$ on each sub-interval
$[t_{k-1},t_k]$ for $1\leq k\leq l+1$.  $\mathcal{S}^{\splineorder}_{\pi}$ is
the set of \emph{splines} of order $\splineorder$ on $\pi$. Elements of $\pi$
are the knots of the splines. Splines have interesting properties (see for
example \cite{DeBoor1993Bsplines}). For instance, they can approximate
arbitrarily well functions in $L^2([a,b])$, provided enough knots are used
(i.e. $l$ is large enough). Moreover, $\mathcal{S}^{\splineorder}_{\pi}$ has a
basis which is made of $l+\splineorder$ functions called the B-splines of
order $\splineorder$ on $\pi$ and denoted $B^{\splineorder}_{k,\pi}$ for
$1\leq k\leq l+\splineorder$. B-splines are easy to calculate, have local
support and have very good numerical properties: finding coordinates of
projected functions with equation \ref{eqDistortionBasis} is both fast and
accurate.

The choice of B-splines as basis functions leads to
$\mathcal{A}=span(B^{\splineorder}_{1,\pi},\ldots,B^{\splineorder}_{l+v,\pi})=\mathcal{S}^{\splineorder}_{\pi}$.
As for truncated Hilbert bases, $l$ can be automatically chosen by a
leave-one-out method.  The choice of $\splineorder$ is more complex as it
corresponds to a regularity assumption.  The most common choice is
$\splineorder=4$, which corresponds to $C^2$ functions. In some situations, it
is interesting to work with derivatives of the original functional data, in
which case higher values of $\splineorder$ will be more adapted. The last
point to address is the choice of $\pi$.  While expert prior knowledge can
justify irregular positioning of the knots (for instance more knots in rough
parts of the studied functions), in general regular knot positions are used,
i.e. $t_k=a+k\frac{b-a}{l+1}$.

Of course, this section only presented some of the possible choices for the
basis. We refer the interested reader to \cite{RamseySilverman97}, especially
to chapters 3, 4 and 15, for a more detailed discussion on function
representations.

\subsubsection{Leave-one-out}\label{sectionCrossValidation}
While expert knowledge or practical considerations can help to choose the
basis among several possibilities such as B-splines or truncated Hilbert
bases, the ideal projection space cannot in general be fixed \emph{a
  priori}. A simple solution is to rely on leave-one-out to compare
different function approximation models.

Let us consider the situation in which we have to compare two candidate
projection sub-spaces, $\mathcal{A}=span(\phi_1,\ldots,\phi_q)$ and
$\mathcal{B}=span(\psi_1,\ldots,\psi_l)$. Let us consider for now only one
function $g$ given by the list $(x_j,y_j)_{1\leq j\leq m}$. We denote
\[
\tilde{g}(x_j,\mathcal{A})=\sum_{k=1}^q\alpha_k\phi_k(x_j),
\]
where $\alpha_k$ is the  $k$-th coordinate of the optimal projection of $g$ in
$\mathcal{A}$ determined by minimizing equation \ref{eqDistortionBasis}.
Similarly, we denote
\[
\tilde{g}(x_j,\mathcal{B})=\sum_{k=1}^l\gamma_k\psi_k(x_j),
\]
where $\gamma_k$ is the $k$-th coordinate of the optimal projection of $g$ in
$\mathcal{B}$ determined by minimizing equation \ref{eqDistortionBasis}
adapted to the $(\psi_k)_{1\leq k\leq l}$ basis.  Comparing the reconstruction
errors given by equation \ref{eqDistortionBasis} is not relevant, as it leads
to overfitting.  Instead, a leave-one-out estimate of the reconstruction
errors of both models is preferred. More precisely, the $\alpha_k^{(-p)}$ are
defined as the optimal coefficients found when one observation is removed from
the list that defines the considered functional data.  The $\alpha_k^{(-p)}$
thus minimize
\[
\sum_{j=1,j\neq
  p}^{m}\left(y_j-\sum_{k=1}^q\alpha_k^{(-p)}\phi_k(x_j)\right)^2.
\]
The $\beta_k^{(-p)}$ coefficients are defined in a similar way on the
$\mathcal{B}$ subspace, using the $\psi_k$ basis instead of the $\phi_k$ one.
The leave-one-out score associated to $\mathcal{A}$ is then given by:
\[
LOO(g,\mathcal{A})=\frac{1}{m}\sum_{i=1}^m\left(y_i-\sum_{k=1}^q\alpha_k^{(-i)}\phi_k(x_i)
\right)^2,
\]
and similarly for $\mathcal{B}$:
\[
LOO(g,\mathcal{B})=\frac{1}{m}\sum_{i=1}^m\left(y_i-\sum_{k=1}^l\gamma_k^{(-i)}\psi_k(x_i)
\right)^2.
\]

In our functional data context, we do not have only one function $g$ given by
the list $(x_j,y_j)_{1\leq j\leq m}$, but a set of $g^i$ functions known
through $(x^i_j,y^i_j)_{1\leq j\leq m^i}$.  In that case, in order to
choose between the $\mathcal{A}$ and $\mathcal{B}$ projection subspaces, the
total cross-validation scores that are obtained by summing the leave-one-out
scores obtained on each function $g^i$ must be compared.

In general, leave-one-out is a very computationally-intensive operation.
Fortunately, this is not the case with linear representations, as the
expansions on $\mathcal{A}$ and $\mathcal{B}$ chosen in the previous section.
Indeed, as the optimization problem of equation \ref{eqDistortionBasis} is
quadratic, there is a matrix $S(\mathcal{A})$ (resp. $S(\mathcal{B})$) such
that $\tilde{g}(x,\mathcal{A})=S(\mathcal{A})y$ (resp.
$\tilde{g}(x,\mathcal{B})=S(\mathcal{B})y$), where
$\tilde{g}(x,\mathcal{A})=(\tilde{g}(x_1,\mathcal{A}),\ldots,\tilde{g}(x_m,\mathcal{A}))^T$
and $y=(y_1,\ldots,y_m)^T$. Then, we have (see \cite{RamseySilverman97}
chapter 10 for instance):
\[
LOO(g,\mathcal{A})=\frac{1}{m}\sum_{i=1}^m\left(\frac{y_i-\tilde{g}(x_i,\mathcal{A})}{1-S(\mathcal{A})_{ii}}\right)^2
\]
A similar equation is satisfied by $LOO(g,\mathcal{B})$.  In general, the calculation of $S$ is much more efficient than the direct
calculation of the cross-validation score.

\subsection{Functional principal component analysis}\label{SectionFunctionalPCA}
Even if the representation on a subspace allows to take into account irregular
sampling and very high original input dimensions, it happens frequently in
practice that a rather high number of basis functions has to be used to keep a
good accuracy for the input function reconstructions. Unfortunately, many data
analysis methods suffer from the curse of dimensionality and are therefore not
really adapted to a high number of input features. In traditional numerical
settings, a simple solution consists in working on a few principal components.

Principal Component Analysis (PCA) was one of the first data analysis methods
adapted to functional data (see
\cite{DauxoisPousse76,DauxoisPousseRomain82,BesseRamsay1986}). On a
theoretical point of view functional PCA consists, as traditional PCA, in
finding an optimal subspace representation. Given $n$ functions
$g^1,\ldots,g^n$ in $L^2(V)$, their $q$ principal functions are defined as $q$
orthonormal functions $\xi_1,\ldots,\xi_q$ such that the following distortion
is minimized:
\begin{equation}\label{eqPCAFunc}
\sum_{i=1}^n\left\|g^i-\sum_{k=1}^q\langle g^i,\xi_k\rangle\xi_k\right\|^2.
\end{equation}
As explained previously, an exact implementation of such a minimization is not
possible: the functions $g^i$ are not known and exact calculation of inner
products and other elementary operations is difficult. A possible solution is
to apply the general method exposed in section \ref{subsectionSubSpace}, i.e.
to work in a subspace $\mathcal{A}$. In this context, as demonstrated in
\cite{RamseySilverman97} (chapter 6), it appears that functional PCA can be
implemented by performing a classical PCA on the transformed coordinates (the
$\beta(\tilde{g}^i)=U\alpha(\tilde{g}^i)$ vectors in \R{q}, see
\ref{sectionCoeff}) of the studied functions. 

This method produces principal vectors in \R{q} that can be transformed back
into principal functions in $\mathcal{A}$. For instance if $\tau$ is such a
vector and $U$ is the Choleski factor of $\Phi$ defined in \ref{sectionCoeff},
then $U^{-1}\tau$ gives the coordinates of the corresponding principal
function $\xi$ on the chosen basis for $\mathcal{A}$. Coordinates of the
original functions on the principal function basis are obtained through inner
products in $\mathcal{A}$.  In practice they are obtained by canonical scalar
products in \R{q} between $\beta(\tilde{g}^i)$ and $\tau$.  Note that unlike
conventional PCA, functional PCA usually works on centered, but not reduced to
unit variance, data, because functional data must be seen as a unique entity
rather than a set of unrelated coordinates with individual scales.

\subsection{Functional transformation}\label{sectionTransformation}
A very interesting aspect of FDA is the possibility to implement a functional
transformation before the data analysis phase. We will not cover in this paper
registering techniques that allow to get rid of time shifting and other
problems that can be interpreted as noise or distortion in the measurement
process, i.e. problems related to the $x^i_j$. We refer the reader to
\cite{RamseySilverman97} chapter five for an introduction on this complex
topic.

We focus here on simpler functional transformations that provide different
views of the same data. For instance, it is quite common in FDA to focus on
the shape of the functions rather than on the actual values. A simple way to
do this is to center and scale functions on a functional point of view, that
is function by function. More precisely, we center $g$ by replacing it by
$g_c$ defined as:
\[
g_c(x)=g(x)-\frac{1}{|V|}\int_Vg(x)\d x.
\]
In this equation, $|V|$ is the volume of the compact $V$ (i.e. $|V|=\int_V\d
x$). The centered function is then scaled into $g_s$ defined as:
\[
g_s(x)= \frac{g_c(x)}{\frac{1}{|V|}\left\|g_c\right\|}.
\]
An interesting aspect of those transformations is that they are based on
elementary operations in the considered functional space. Therefore, they can
be implemented using the coefficients that represent the input functions
on the chosen projection space, as explained in section \ref{sectionCoeff}.

Another way to focus on shapes rather than on values is to calculate
derivatives of the considered functions. To do so, we have to choose a
projection space with a basis formed by derivable functions. Then, if
$\tilde{g}^i=\sum_{k=1}^q\alpha_k(\tilde{g}^i)\phi_k$, obviously
$\tilde{g}^{i(s)}=\sum_{k=1}^q\alpha_k(\tilde{g}^i)\phi_k^{(s)}$, where
$f^{(s)}$ corresponds to the $s$-th derivative of $f$. It is therefore
possible to work in $\mathcal{A}^{(s)}=span(\phi_1^{(s)},\ldots,\phi_q^{(s)})$
exactly as we did in $\mathcal{A}$ (note that
$(\phi_1^{(s)},\ldots,\phi_q^{(s)})$ might not be a free system anymore).

The special case of B-spline bases is very interesting. Indeed, it is clear
that the derivative of a spline from $S^{\splineorder}_{\pi}$ is a spline of
$S^{\splineorder-1}_{\pi}$; therefore it uses the same knots. Moreover, the
coordinates of the derivative spline on the order $\splineorder-1$ B-spline
basis can be calculated exactly and very easily with a finite difference
equation, using the coordinates of the original spline on the order
$\splineorder$ B-spline basis (see \cite{DeBoor1993Bsplines}).  This allows to
work with derivatives exactly as with the original functions.

\section{Simulation results}\label{sectionSimulations}

The functional approach to Radial-Basis Function Networks and Multi-Layer
Perceptrons is illustrated on spectrometric data coming from the food
industry.  This benchmark has been chosen here for illustration purposes: it
permits to show which kind of functional (pre)processing is expected to give
results similar to those on the original data, and which ones could lead to
improved results.  The models (RBFN and MLP) are optimized as detailed below.
Nevertheless, their learning algorithm is chosen a priori, and no attempt is
made to improve the results by comparing to other learning strategies; only
comparisons between the possible ways to handle the functional data are
discussed here.

\subsection{Tecator spectra benchmark}

The Tecator data set \cite{TecatorDataSet} consists of 215 near-infrared
absorbance spectra of meat samples, recorded on a Tecator Infratec Food and
Feed Analyzer.  Each observation consists in a 100-channel absorbance spectrum
in the 850-1050 nm wavelength range. Each spectrum in the database is
associated to a content description of the meat sample, obtained by analytic
chemistry; the percentage of fat, water and protein are reported.  The goal of
the benchmark here is to predict the fat percentage from the spectra; this
percentage is in the $[0.9, 49.1]$ range.

From the 215 100-dimensional spectra, 43 are kept aside as a test
set; the test set will not be used neither for model learning, nor
for cross-validation (selection of the number of splines, of the
number of PCA components, of the number of parameters in the models,
etc.). The 172 remaining samples are used for model learning and
validation, as detailed below.

\begin{figure}[htbp]
\centering
\includegraphics[width=\textwidth]{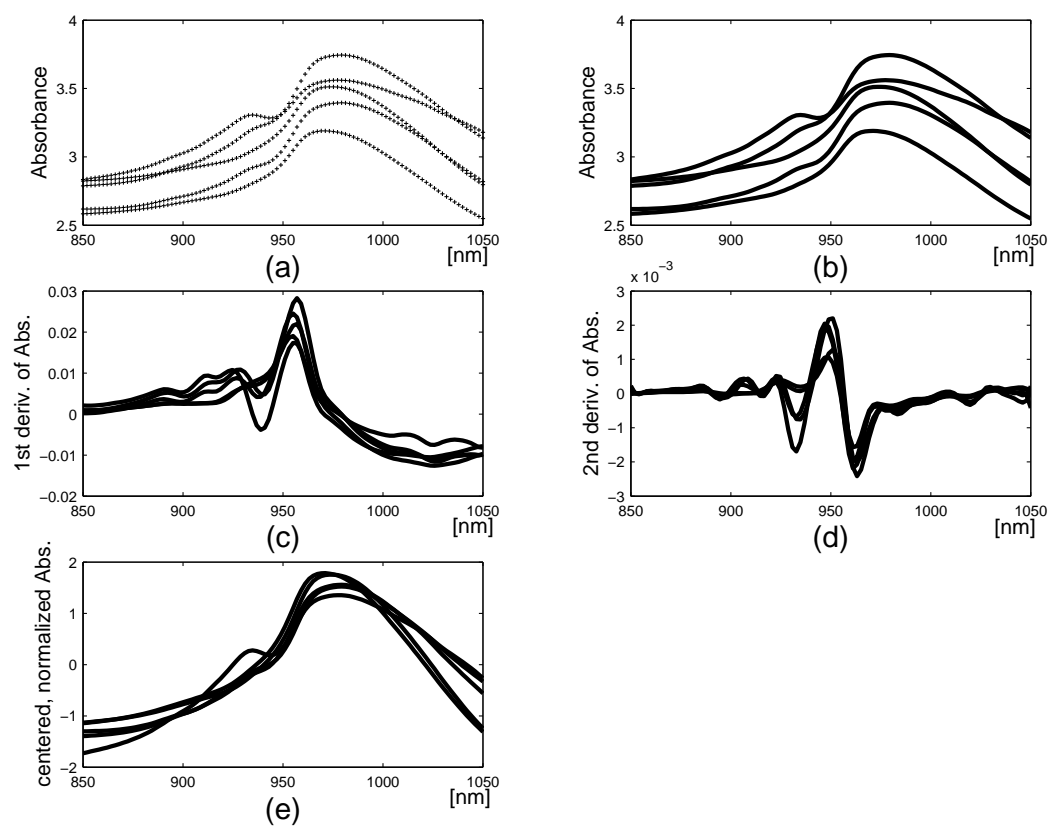}
\caption{Five spectra from the Tecator benchmark; (a) original samples, (b) 4th-order spline approximation, (c) derivative of 5th-order spline approximation, (d) second derivative of 6th-order spline approximation, (e) 4th-order spline approximation after functional centering and reduction}
\label{fig_Tecator_preProcessRbfn}
\end{figure}

It should be mentioned that the spectra are finely sampled, leading to very
smooth curves; some of them are illustrated in
Figure~\ref{fig_Tecator_preProcessRbfn}(a). It is therefore not expected that
a functional preprocessing of the rough data, such as a spline decomposition,
will lead to improved results, except if some a priori information is added,
or if irregular sampling is artificially created by omitting data. This last
point will be detailed in section \ref{sectionMissingData}.

\subsection{Preprocessing the Tecator spectra}\label{subsectionPreprocessTecator}

In addition to working with rough spectra (100-dimensional measured
vectors), three types of preprocessing are considered in the
experiments.

\begin{figure}[htbp]
\centering
\includegraphics[width=\textwidth]{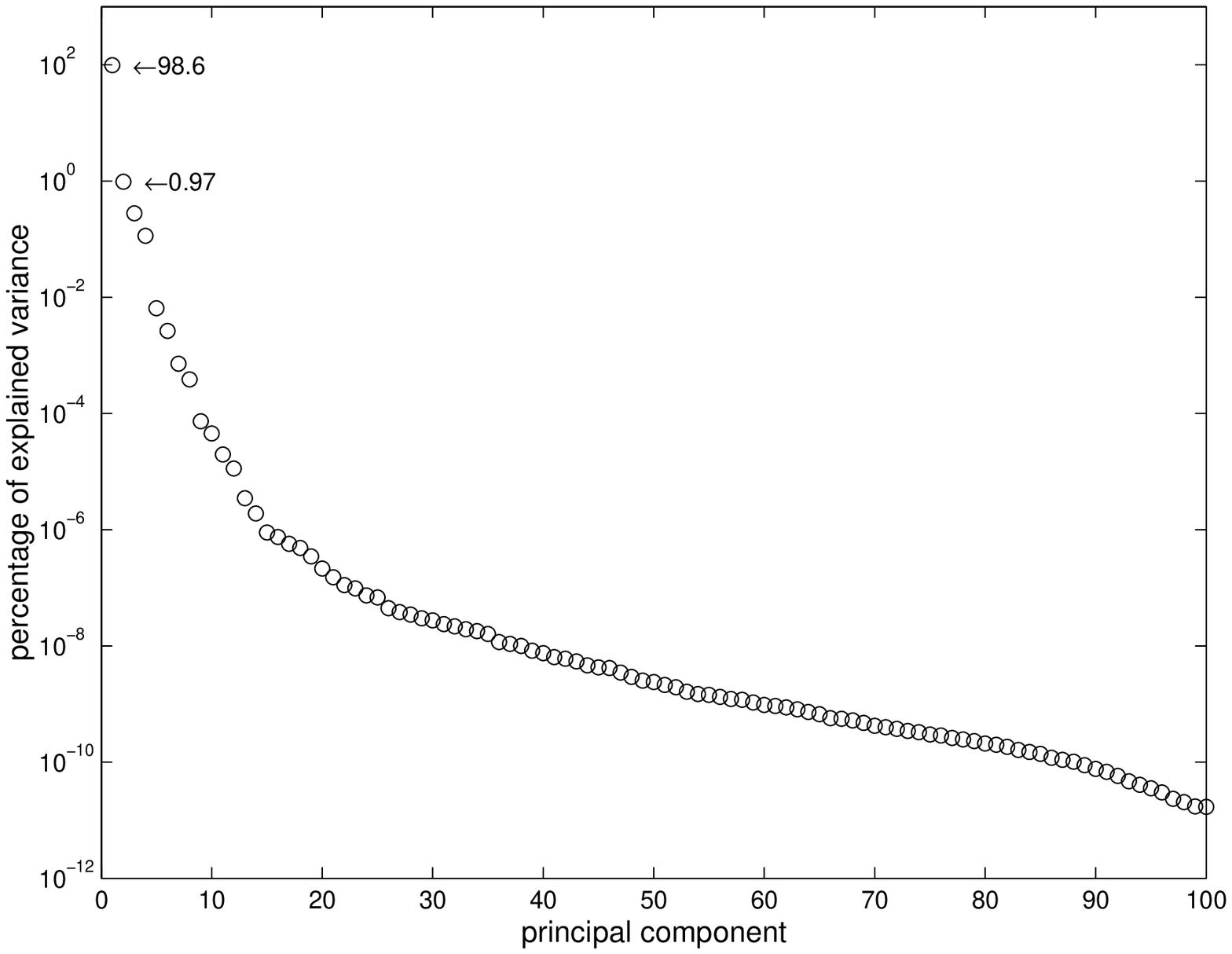}
\caption{Percentage of variance explained by each of the PCA principal components}
\label{fig_Tecator_PCA}
\end{figure}

\begin{figure}[htbp]
\centering
\includegraphics[width=0.8\textwidth]{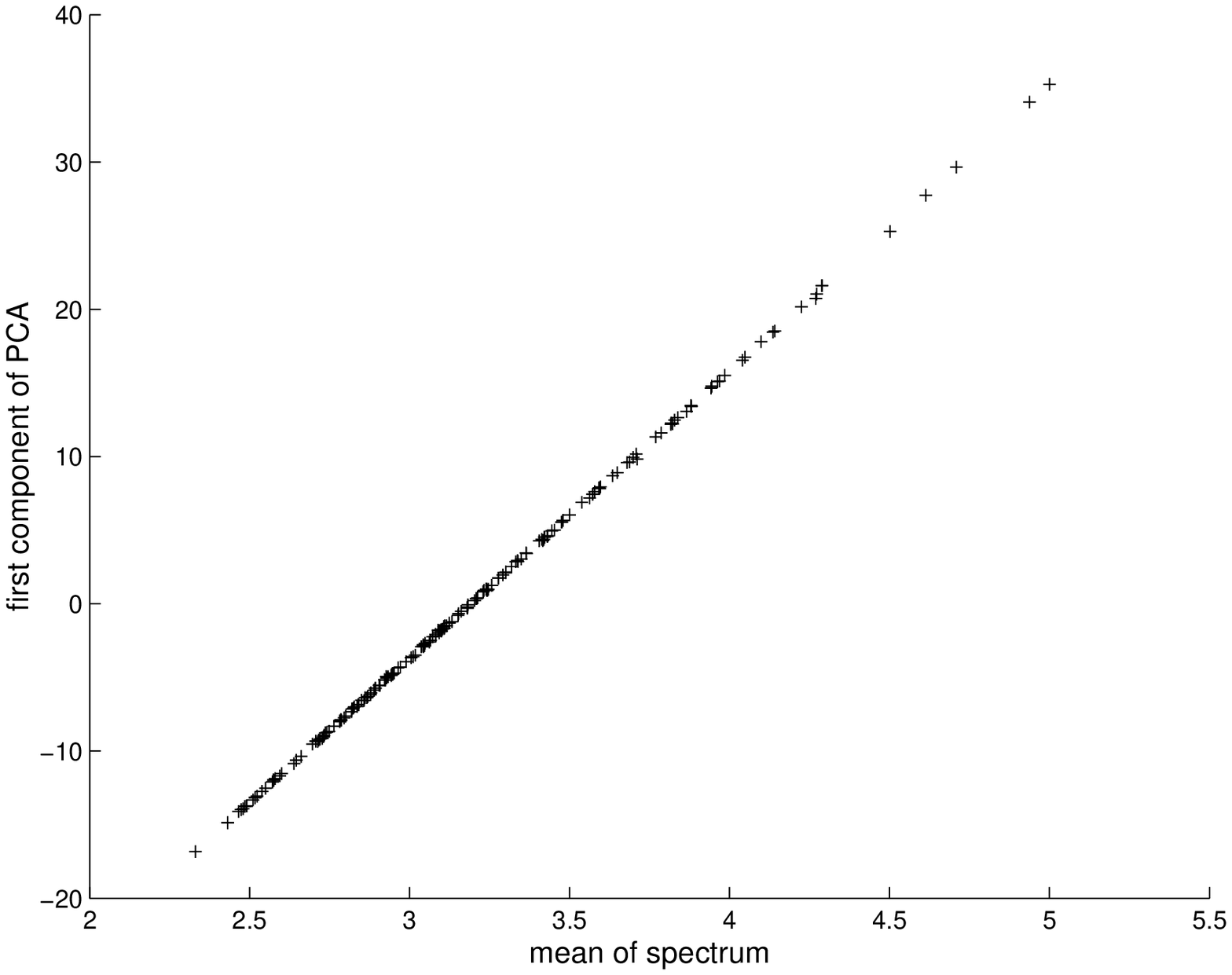}
\caption{Coordinates on the first principal component as a function of the
  spectrum mean}
\label{figurePCAMean}
\end{figure}

First of all, a standard Principal Component Analysis (PCA) is performed.
Unsurprisingly because of the smooth character of spectra, most of the
information in terms of percentage of variance is contained in a few PCA
components; this is illustrated in Figure~\ref{fig_Tecator_PCA} that shows the
percentage of variance of the original data associated to each eigenvalue.  As
usually, data have been centered and reduced before applying the PCA; this
induces a scaling that may have influence on the inner products and distance
computations.  The PCA is a non-functional preprocessing, as the continuous
structure of spectra is not taken into account.  It can be seen easily that
the first principal component almost exactly represents the spectrum means, as
illustrated by figure \ref{figurePCAMean}.

To take into account the smooth character of spectra, a functional
preprocessing is performed using bases of splines.  Splines of
degrees 3, 4 and 5 (respectively order 4, 5 and 6) are used; the two
last ones are aimed to be derived (respectively once and twice) in
order to work with derivatives of spectra rather than with the original
ones.

The numbers of basis functions selected by the leave-one-out procedure
detailed in section~\ref{sectionCrossValidation} are respectively 48, 43 and
32 for the 4th-, 5th- and 6th-order splines.  Figures
~\ref{fig_Tecator_preProcessRbfn}(b), (c) and (d) respectively show the
spectra approximated by 4th-order spline, the derivative of the 5th-order
spline approximation, and the second derivative of the 6th-order spline
approximation.

A last preprocessing takes into account a priori information on the spectra.
As spectrometry experts know that the shape of the spectra is by far more
important than their mean value (for the fat prediction problem), centering
and reducing them (to unit variance) avoids that their average could influence
the models.  Figure~\ref{fig_Tecator_preProcessRbfn}(e) shows the results of
this functional centering and reduction (performed on the 4th-order spline
approximation).

\subsection{Using the Radial-Basis Function Network model}\label{sectionRBFResults}

A number of experiments have been conducted on the Tecator
benchmark, with the Radial-Basis Function Network model described in
section~\ref{sectionRBFN}.  The model parameters are learned through an OLS
procedure, as detailed in~\cite{cit_Chen_OLS}. In short, the OLS procedure is a
forward selection algorithm that incrementally chooses centers (and
associated Gaussian functions) among a set of candidates; here the
candidates are the data in the learning set.  The error criterion
used for the selection is the sum of the squared errors made by the
model and a regularization term~\cite{cit_Orr_RegRbfn}; the contribution of each candidate
to the error criterion is measured, and the one that minimizes this
contribution is chosen.  The incremental procedure is performed up
to a high number of selected Gaussian functions (100 in the
following experiments).  Next, a 4-fold cross-validation procedure
is used on the learning set (172 spectra) to select the optimal
number of Gaussian functions, according to the sum-of-squares error
criterion.

Ten experiments are performed:
\begin{enumerate}
\item The 100-dimensional rough spectra are used as inputs to the
RBFN.
\item The spectra are preprocessed by PCA, and the first 20 PCA components are
  kept (100\% of the original variance is preserved, as shown in Figure
  \ref{fig_Tecator_PCA}).
\item The same PCA preprocessing is used, but now the number of PCA components
  that are kept is selected according to a 4-fold cross-validation procedure
  on the learning set; this optimization leads to a choice of 5 PCA
  components.
\item The PCA components are used, but they are whitened (centered and scaled
  to unit variance); unlike the functional centering and reduction detailed in
  the previous section, the whitening here is a conventional one, i.e. it is
  applied component by component. The purpose of this whitening is to allow
  each component to have the same importance.  Without this whitening, the
  first component would have much more importance than the other ones, while
  its influence on the fat prediction problem is known to be low.  The number
  of PCA components that are kept is selected by 4-fold cross-validation as
  above: 6 components are selected.
\item The 48 coefficients of the 48 4th-order splines are used as
inputs to the RBFN.
\item A functional PCA (see Section \ref{SectionFunctionalPCA}) is performed
  on the 4th-order splines with 48 coefficients; 
  20 coefficients are kept.
\item A functional PCA is performed on the 4th-order splines with 48
  coefficients, and the PCA coefficients are whitened; A 4-fold cross-validation selects 6 components.
\item A functional centering and reduction is applied to the
4th-order spline approximation; the 48 resulting coefficients are
used as inputs to the RBFN.
\item The 5th-order spline is derived, and the 42 resulting
coefficients are used as inputs to the RBFN.
\item The 6th-order spline is derived twice, and the 30 resulting
coefficients are used as inputs to the RBFN.
\end{enumerate}

Table \ref{resultsRBFN} shows the results of these ten
experiments. All results are given in terms of Root Mean Square
Error (RMSE) on the test set.

\begin{table}[htbp]
  \begin{center}
\begin{tabular}[\textwidth]{|c|p{16em}|c|}

  \hline
  Experiment \# & Experiment & Result on test set \\
  \hline
  1 & 100-dimensional original data & 4.97 \\
  2 & PCA, no whitening, 20 components & 4.99 \\
  3 & PCA, no whitening, a 4-fold cross-validation selects the 5 first components & 4.85 \\
  4 & PCA, whitening, a 4-fold cross-validation selects the 5 first components & 1.94 \\
  5 & 4th-order B-splines, 48 coefficients, no whitening & 4.59 \\
  6 & 4th-order B-splines, 48 coefficients, functional PCA, no whitening, 20 components & 4.59 \\
  7 & 4th-order B-splines, 48 coefficients, functional PCA, whitening, a 4-fold cross-validation selects the 6 first components & 1.83 \\
  8 & 4th-order B-splines after functional centering and reduction, 48 coefficients, no whitening & 1.64 \\
  9 & first derivative of 5th-order B-splines, 42 coefficients, no whitening & 0.90 \\
  10 & second derivative of 6th-order B-splines, 30 coefficients, no whitening & 0.81 \\
  \hline
\end{tabular}
\caption{RMSE on the test set for the RFBN experiments (see text for details)}\label{resultsRBFN}
\end{center}
\end{table}

The following conclusions can be drawn.
\begin{itemize}
\item The results of experiments 1 and 5 are roughly the same.  Indeed the
  decomposition into splines does not bring any improvement, as there is a
  nearly perfect correspondence between the original spectra and their spline
  approximation.  The use of the scaling after Cholesky decomposition of the
  $\Phi$ matrix (see section~\ref{subsectionSubSpace}) guarantees that the
  results obtained after spline preprocessing will be similar to those on the
  original spectra, as the latter are very smooth (therefore almost perfectly
  approximated by splines).
\item The non-functional PCA reduction (experiments 2 and 3) does not bring
  any improvement to the results; the reduction to unit variance included in
  the PCA does not seem to be advantageous here; actually the variances of the
  original data components are more or less identical in the data set,
  therefore the reduction has little effect.
\item The centering and reduction of the PCA coefficients (experiment 4)
  improves the results; indeed the influence of the first PCA component is
  strongly decreased in this process.  The first PCA component is proportional
  to the spectrum averages (see Figure \ref{figurePCAMean}), which are known to
  be of little influence in the fat prediction problem.
\item Similarly to the fact that experiments 1 and 5 give approximately the
  same results (the decomposition into splines does not bring much additional
  smoothness as the original spectra are already very smooth), experiments 2
  and 6 on one side, and experiments 4 and 7 on the other side, lead to
  similar results.  In experiments 2 and 4 an initial reduction to unit
  variance is performed and not in experiments 6 and 7, but as mentioned above
  this reduction does not bring any improvement.
\item The functional centering and reduction (experiment 8) also improves the
  predictions compared to the original ones.  The improvement also results
  from the removal of the spectrum averages, and is comparable to the
  centering and reduction of the PCA components.
\item As expected, taking the first and second derivatives of the spectra
  (more precisely, taking the first and second derivative of their 5th- and
  6th- order spline approximations respectively) focuses on the differences in
  the spectra shapes, therefore allowing a better prediction of fat content.
\end{itemize}

\subsection{Using the Functional Multi-layer Perceptron model}\label{subsectionResultsMLP}
The original contributor of the Tecator data set used traditional MLP together
with PCA to build a regression model
\cite{BorggaardThodberg1992,Thodberg1996}. In \cite{BorggaardThodberg1992}
Borggaard and Thodberg use a standard MLP on the 10 first principal
components. They use early stopping to avoid overfitting and report a RMSE of
0.65. In \cite{Thodberg1996} Thodberg reports better results based on a more
complex training algorithm and model: he uses a weight decay regularization
term and chooses meta-parameters through a Bayesian approach. More
precisely, he uses an one hidden layer perceptron with additional direct
connections from the inputs to the output node (introducing this way a linear
term) and three separate regularization terms, one for each layer and one for
the weights of the linear term. The values of those weight decays as well as
the number of hidden neurons are determined by a Bayesian estimation of the
generalization error. Using the 10 first principal components Thodberg obtains
a RMSE of 0.55. In order to improve the results, he combines the 10 best MLPs
obtained out of 40 trained MLPs in an ensemble model that reaches a RMSE of
0.52. Finally, he embeds into the Bayesian meta-parameters selection the
determination of the optimal number of principal components. He selects this
way 12 principal components for a RMSE of 0.42. Using a smoothed version of
this input selection (based on two other weight decay parameters) he even
managed to reach a RMSE of 0.36 with 13 principal components.

The goal of the proposed simulations is not to reproduce Thodberg's results
but simply to illustrate the positive effects of the functional methodology.
Therefore, a simplified neural model has been used in order to focus on
the preprocessing. The chosen model is a single classical one hidden layer
perceptron with no direct connection, together with a single regularization
term (which is not used for bias terms). Meta-parameters (the weight decay,
the number of hidden neurons, etc.) are chosen through the same 4-fold
cross-validation procedure (as with the RBFN models). No ensemble model or
smooth variable selection is used. Training itself is done by a second-order
gradient descent method starting from 60 different initial random weight
vectors (for each experiment). The best MLP obtained from those random weight
is kept according to the sum-of-squares error criterion (on the training set)
combined with the regularization term. The number of hidden neurons varies
from 1 to 6.  In all experiments, the best 4-fold cross-validation was
obtained with 2 hidden units.

Five experiments were conducted:
\begin{enumerate}
\item The spectra are preprocessed by PCA and the number of PCA components is
  selected by the 4-fold cross-validation procedure on the learning set. This
  experiment plays the role of the reference one as we do not use the
  sophisticated method of Thodberg.
\item The spectra are converted into their 4th-order B-splines representation,
  a functional PCA is conducted and the number of components to retain is
  again selected by the 4-fold cross-validation procedure.
\item We do the same as in the previous experiment but we apply a functional
  centering and reduction before the functional PCA.
\item The 5th-order spline is derived; a functional PCA is conducted on the
  resulting functions and the number of components to retain is
  again selected by the 4-fold cross-validation procedure.
\item The 6th-order spline is derived twice; functional PCA is conducted on the
  resulting functions and the number of components to retain is
  again selected by the 4-fold cross-validation procedure.
\end{enumerate}
An important point is that PCA coordinates are always whitened before being
used by the MLP. Indeed as the explained variance is very much
concentrated in the first coordinate, the variation range is quite different
from the different inputs of the MLP. This means that the corresponding
weights should be very different: this is not compatible with the weight decay
regularization as large weights (corresponding to large range) are more heavily
penalized than small weights.

Another difference with the RBFN experiments is that a PCA (functional or
classical) is always done before submitting the data to the MLP. Our goal was
to avoid huge training times as well as high dimensionality related problems
induced by the size of the data. The experiments have been limited this way to
at most 18 principal components to use a reasonable input size for the MLP.

Table \ref{tableResultsMLP} summarizes the results of those five experiments.
All results are given in terms of Root Mean Square Error (RMSE) on the test
set.

\begin{table}[htbp]
  \centering
  \begin{tabular}{|c|p{16em}|c|}\hline
  Experiment \# & Experiment & Result on test set \\\hline
 1 & PCA, a 4-fold cross-validation selects the 12 first components, whitening & 0.49\\
 2 & 4th-order B-splines, 48 coefficients, functional PCA, a 4-fold cross-validation selects the 12 first components, whitening & 0.49\\
 3 & 4th-order B-splines, 48 coefficients, functional PCA on centered and reduced functions, a 4-fold cross-validation selects the 11 first components, whitening & 0.44\\
 4 & first derivative of 5th-order B-splines, 42 coefficients, functional PCA, a 4-fold cross-validation selects the 15 first components, whitening & 0.50 \\
 5 & second derivative of 6th-order B-splines, 30 coefficients, functional PCA, a 4-fold cross-validation selects the 13 first components, whitening & 0.61\\\hline
  \end{tabular}
  \caption{RMSE on the test set for the MLP experiments}
  \label{tableResultsMLP}
\end{table}

These results justify the following comments.
\begin{itemize}
\item The reference experiment (number 1) shows that the chosen experimental
  setting is comparable to the one used by Thodberg. Indeed, the obtained MLP
  performs slightly better than the one selected by Thodberg (0.49 versus
  0.55) probably because we choose automatically the appropriate number of
  principal components. On the other hand, our simpler setting cannot reach
  the best performances reported in \cite{Thodberg1996} probably because the
  regularization method is less flexible than the one used by Thodberg.
\item The best functional preprocessing allows to improve slightly the test
  performances (from 0.49 to 0.44, that is about 10 \%) in a rather simple
  way. Moreover, a higher-level 4-fold cross-validation in which the method
  itself is automatically chosen in addition to the number of hidden neurons,
  of principal components, etc., chooses the model produced by experiment
  number 3, i.e. the functional preprocessing that leads to the best test
  performances.
\item MLP performances are better than RBFN ones, but the price to pay in
  terms of calculation time is huge. A full experiment with MLPs takes about
  200 times more computational time than a similar experiment with a RBFN
  network. We face here one of the classical tradeoffs between model design
  time and model accuracy. As the functional approach introduces additional
  possibilities such as functional preprocessing (derivative, centering,
  etc.), it makes the training problem even more important. Exploring all the
  available functional preprocessing solutions can become nearly impossible
  for MLP models while remaining feasible for RBFN networks. Moreover, it
  appears clearly that RBFN results cannot be used as a guideline for the
  construction of a good MLP model. Indeed, derivatives were really useful for
  improving the RBFN results, whereas they give worse performances in the case
  of the MLP.
\end{itemize}

\section{Missing data}\label{sectionMissingData}
\subsection{A semi-artificial benchmark}
A nice property of FDA is its ability to deal with irregular sampling. In some
situations, it happens that the sampling process has some variation between
input functions. This is the case for instance in medical time series where
patients decide on their own when to be monitored by doctors. Irregular
sampling appears also for gesture recognition like cursive handwriting
recognition for personal digital assistant: while the sampling rate is fixed,
gestures have different execution times that depend on the context of
execution rather than on the gesture performed. Therefore, some registration
is needed; its effect is to transform the regular sampling into a
gesture-specific one.

The goal of this section is to illustrate the way FDA solves irregular
sampling problems in its simplest form: a regular sampling with missing data.
To do so, a semi-artificial data set was created by removing data from the
Tecator data set used in the previous section. More precisely, 10 \% of the
observations in each spectrum of the data set were removed at random
(therefore 90 absorbances out of 100 are kept). Of course, spectrometers
provide regular spectra and the obtained data are not representative of
spectrometric problems. The goal is simply here to illustrate the
possibilities of FDA with data for which we have reference performances.

\subsection{Functional preprocessing}
The function representation strategy described in section
\ref{sectionFunctionRepresentation} applies to arbitrary sampling.  Therefore
the procedure followed in section \ref{subsectionPreprocessTecator} does not
have to be modified: the coordinates of the considered functions on B-spline
bases of various orders are calculated. The leave-one-out procedure selects
less B-splines than with complete data. Indeed, the number of basis functions
are respectively 28, 27 and 21 for the 4th-, 5th- and 6th-order splines. This
reduction is easily explained by the fact that B-splines are localized
functions. When the number of knots is high, the support of individual
B-splines is small and it can happen that for a given spectrum no observation
is available on the whole support of a B-spline. In this case, the
corresponding coordinate cannot be calculated. Before this extreme situation,
coefficients become numerically unstable because some B-spline supports do not
contain enough observations to allow a correct estimation of the corresponding
coefficients.

\subsection{Results}

The Tecator benchmark with 10 \% of missing data as described above was used
for experiments with a RBFN network.  Only the most interesting methods from
section \ref{sectionRBFResults} were used on these data, namely experiments
number 9 and 10, on the first and second derivative of the 5th- and 6th-order
splines respectively.  Table \ref{tableResultsRBFMissing} gives the results of
the two experiments (to simplify the comparison with section
\ref{sectionRBFResults}, the same experiment numbers are kept). It clearly
appears that the functional approach solves the problem of missing data in
this particular situation.

\begin{table}[htbp]
  \centering
  \begin{tabular}{|c|p{16em}|c|}\hline
  Experiment \# & Experiment & Result on test set \\\hline
 9 & first derivative of the 5th-order B-splines, 26 coefficients, no
 whitening  & 1.05\\
 10 & second derivative of the 6th-order B-splines, 19 coefficients, no whitening  & 0.80\\ \hline
  \end{tabular}
  \caption{RMSE on the test set for the RBF experiments with missing data}
  \label{tableResultsRBFMissing}
\end{table}

As for the RBFN, experiments were limited to the best preprocessing methods
for use with the MLP, namely raw functional data followed by a functional PCA as well as
centered and reduced functional data also followed by a functional PCA. All
meta-parameters (including the number of principal components) were selected
by a 4-fold cross-validation, exactly as in section \ref{subsectionResultsMLP}. Table \ref{tableResultsMLPMissing} gives the
results of the two experiments. It also
clearly appears  that the functional approach solves here the problem of missing
data.

\begin{table}[htbp]
  \centering
  \begin{tabular}{|c|p{16em}|c|}\hline
  Experiment \# & Experiment & Result on test set \\\hline
 2 & 4th-order B-splines, 28 coefficients, functional PCA, a 4-fold cross-validation selects the 12 first
 components, whitening & 0.52\\
 3 & 4th-order B-splines, 28 coefficients, functional PCA on centered and reduced functions, a 4-fold
 cross-validation selects the 11 first 
 components, whitening & 0.44\\ \hline
  \end{tabular}
  \caption{RMSE on the test set for the MLP experiments with missing data}
  \label{tableResultsMLPMissing}
\end{table}

\subsection{Alternative solutions}
The standard way of dealing with missing data is to use an imputation method
that will reconstruct the needed values. The simplest imputation method
consist in replacing a missing value by the mean of available values for the
corresponding variable.

A more interesting method consists in using a $k$-nearest neighbors approach:
given an input with missing values, its $k$-nearest neighbors, among inputs
that do not miss the corresponding value, are calculated and the missing value
is replaced by the mean of this variable for the $k$-nearest neighbors. Of
course the distance has to be adapted to take care of the missing data
problem. A possible solution is simply to discard missing values. Let us
denote $nm(x)$ the set of indices $j$ for which $x_j$ is not missing. Then the
distance used for the nearest neighbor $y$ calculation is:
\[
d(x,y)=\frac{1}{|nm(x)\cap nm(y)|}\sum_{j\in nm(x)\cap nm(y)}(x_j-y_j)^2,
\]
where $|A|$ is the cardinal of the set $A$.

When imputation has been done, a standard processing method can be applied.
Section \ref{sectionRBFResults} showed that non-functional approaches give
very bad results for the RBF network; therefore it has been decided to study
the imputation method associated to a standard processing approach only for
the MLP model. Two experiments have been conducted in this way:
\begin{enumerate}
\item Missing values are imputed using the mean approach to reconstruct
  spectra in \R{100}; then a standard PCA is applied. The number of principal
  components to retain is determined by the 4-fold cross-validation used for
  other meta-parameters optimization.
\item Missing values are imputed using the $k$-nearest neighbors
  approach. Resulting spectra are processed by a regular PCA. Both $k$ and the
  number of principal components are determined by 4-fold cross-validation.
\end{enumerate}
Table \ref{tableResultsMLPMissingImputation} summarizes the obtained results,
which are quite bad, especially for the mean approach. It appears in fact that
the reconstruction is very bad because of the mean spectrum effect already
encountered with the RBFN model. Indeed, spectra with similar shape but very
different means can correspond to similar values of fat. Unfortunately, this
means that reconstructing the shape of the spectra without using an expert
knowledge is very difficult.

\begin{table}[htbp]
  \centering
  \begin{tabular}{|c|p{16em}|c|}\hline
  Experiment \# & Experiment & Result on test set \\\hline
 1 & mean imputation, PCA, a 4-fold cross-validation selects the 5 first
 components,  whitening  & 7.13\\
 2 & 4-nearest neighbors imputation,  PCA, a 4-fold cross-validation selects
 the 12 first components, whitening & 1.87\\ \hline
  \end{tabular}
  \caption{RMSE on the test set for the MLP experiments with missing data and
 classical imputation methods}
  \label{tableResultsMLPMissingImputation}
\end{table}

A possibility to take into account this kind of expert knowledge into the
imputation process without relying on a functional approach is to center and
scale each spectrum before applying the imputation method. More precisely, in a
way modeled after the functional scaling described in section
\ref{sectionTransformation}, $x$ is replaced by $x^s$ defined by:
\[
x^s_i=\frac{x_i-\sum_{j\in nm(x)}x_j}{\sqrt{\sum_{j\in nm(x)}\left(x_j-\sum_{k\in nm(x)}x_k\right)^2}}
\]
The same experiments as described above were conducted with this additional
preprocessing phase. Table \ref{tableResultsMLPMissingFuncImputation}
summarizes the obtained results, which are much better than without the
inclusion of the expert knowledge. Even so, results are still worse than
the ones obtained by the functional preprocessing. Moreover, the expert
knowledge corresponds clearly to a functional point of view and the imputation
methods based on it should be considered as almost functional methods. 

\begin{table}[htbp]
  \centering
  \begin{tabular}{|c|p{16em}|c|}\hline
  Experiment \# & Experiment & Result on test set \\\hline
 1 &  expert pre-processing, mean imputation, PCA, a 4-fold cross-validation
 selects the 9 first components, whitening & 1.82\\
 2 & expert pre-processing, 8-nearest neighbors imputation, PCA, a 4-fold
 cross-validation  selects the 9 first components, whitening & 0.85\\ \hline
  \end{tabular}
  \caption{RMSE on the test set for the MLP experiments with missing data and
 expert imputation methods}
  \label{tableResultsMLPMissingFuncImputation}
\end{table}

\section{Conclusion}
Functional Data Analysis (FDA) is an extension of traditional data analysis to
functional data, lying in an infinitely-dimensional space.
Examples of functional data are spectra, temporal series, spatio-temporal
images, gesture recognition data such as cursive handwriting patterns, etc.
Functional data are rarely known in practice; instead lists of input-output
pairs (one for each functional data) are usually known.  Their sampling can be
irregular, even different from one functional data to another.

This paper shows how to extend the Radial-Basis Function Network (RBFN) and
Multi-Layer Perceptron (MLP) models to functional data inputs.  A particular
emphasis is put on how to handle functional data in practical situations, i.e.
when they are known through list of sampled values.  In particular, various
possibilities for functional processing are presented, including the
projection on smooth bases, Functional Principal Component Analysis (FPCA),
functional centering and reduction and the use of differential operators.  It
is shown how to incorporate these functional preprocessings into the RBFN and
MLP models, and how to take into account the non-orthogonality of basis
vectors in the case of preprocessing by projection.

The methods are applied to a benchmark in spectroscopy.  The advantages and
limitations of the various FDA approaches are discussed on this benchmark,
both in the RBFN and MLP cases.  It is shown how an adequately chosen
functional preprocessing can improve the way functional data are handled into
data analysis methods.

The case of irregularly-sampled functional data is discussed through the same
benchmark where a percentage of values have been artificially removed.  It is
shown that the FDA approach is robust to such missing data, while traditional
imputation techniques fail to provide adequate results.

The FDA approach, combined with an appropriate choice of how to represent
the functional data, may reveal interesting in a variety of situations where
the smooth character or the irregular sampling of data has to be taken into
account.

\bibliographystyle{plain}
\bibliography{total}

\end{document}